\documentclass{article} 
\usepackage{iclr2026_conference,times}

\usepackage{amsmath,amsfonts,bm}

\def\eqref#1{equation~\ref{#1}}

\def\1{\bm{1}}

\DeclareMathAlphabet{\mathsfit}{\encodingdefault}{\sfdefault}{m}{sl}
\SetMathAlphabet{\mathsfit}{bold}{\encodingdefault}{\sfdefault}{bx}{n}

\usepackage{hyperref}
\usepackage{url}
\usepackage{booktabs}
\usepackage{amsfonts}
\usepackage{nicefrac}
\usepackage{microtype}
\usepackage{xcolor}
\usepackage{graphicx}
\usepackage{amsmath}
\usepackage{bm}
\usepackage{booktabs}
\usepackage{multirow}
\usepackage{subfig}
\usepackage{subcaption}

\title{SegSplat: Feed-forward Gaussian Splatting and Open-Set Semantic Segmentation}

\author{Peter Siegel \\
  ETH Z{\"u}rich \\
  \texttt{psiegel@ethz.ch}
  \And
  Federico Tombari \\
  Google \\
  \texttt{dbarath@inf.ethz.ch}
  \And
  Marc Pollefeys \\
  ETH Z{\"u}rich, Microsoft \\
  \texttt{dbarath@inf.ethz.ch}
  \And
  Daniel Barath \\
  ETH Z{\"u}rich \\
  \texttt{dbarath@inf.ethz.ch}
}

\iclrfinalcopy 
\begin{document}

\maketitle

\begin{figure}[h]
  \centering
  \vspace{-8mm}
  \includegraphics[width=\linewidth]{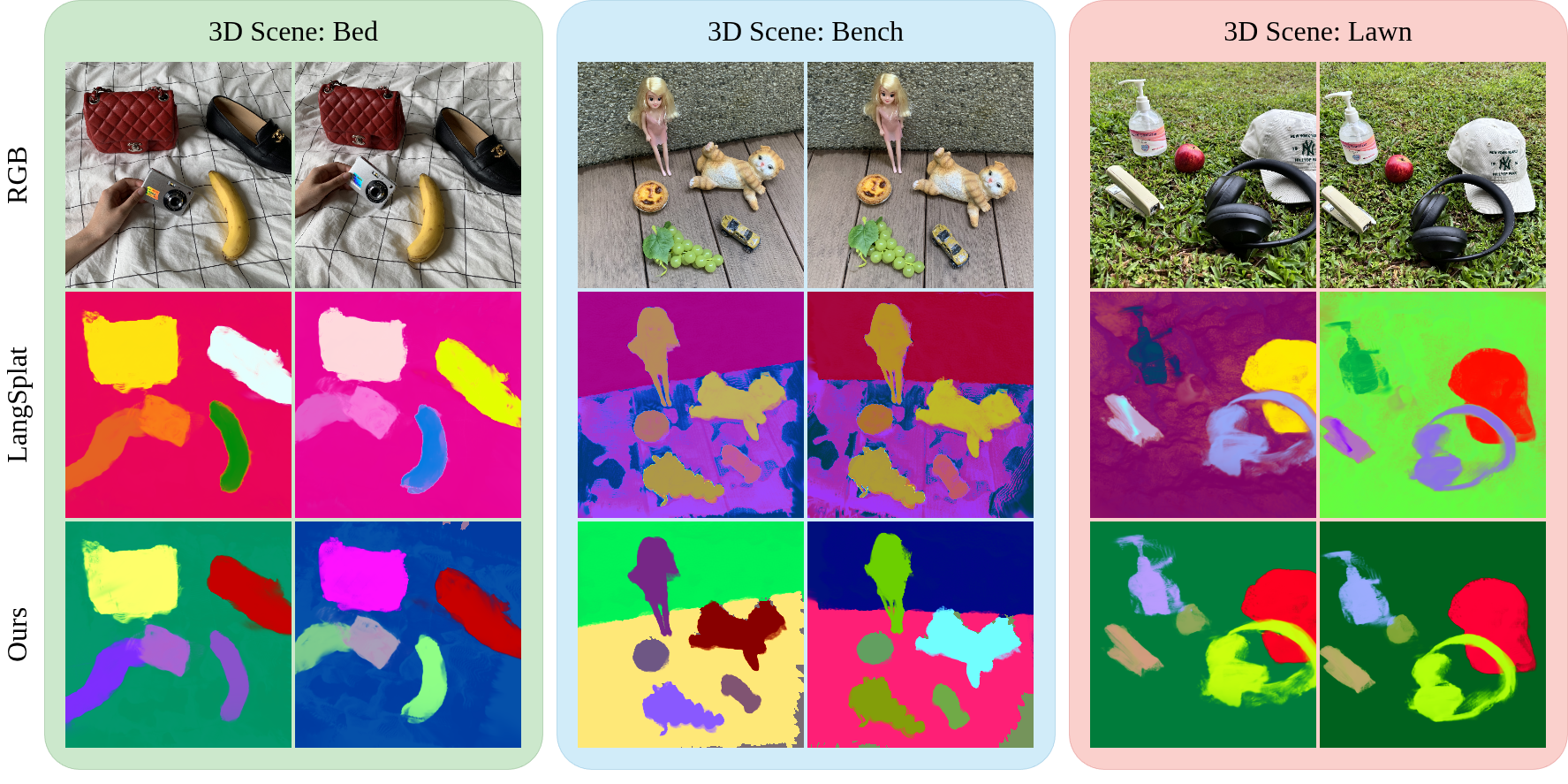}
  \vspace{-2mm}
  \caption{Visualization of learned 3D language features of the previous state-of-the-art method, LangSplat~\cite{qin2024langsplat}, and our SegSplat.
    While LangSplat requires per-scene training and generates imprecise features, our SegSplat captures smooth regions more consistently and needs \textit{no training}.
  While being effective, our SegSplat is also 59$\times$ faster than LangSplat.}
  \label{fig:teaser}
  \vspace{-2mm}
\end{figure}

\begin{abstract}
  Efficient 3D scene reconstruction with open-set semantic understanding is vital for robotics and augmented reality.
  However, existing methods either require costly per-scene optimization to incorporate semantics into expressive 3D representations like 3D Gaussian Splats (3DGS), or, if feed-forward, focus predominantly on geometry and appearance.
  This paper introduces SegSplat, the first framework to predict 3D Gaussian Splats with associated open-set semantic features in a purely feed-forward manner.
  Building upon efficient sparse-view 3DGS techniques, SegSplat uniquely integrates semantic knowledge derived from 2D open-set segmentation models (e.g., SAM and CLIP) without any training.
  We achieve this by constructing a compact semantic memory bank from features in the input images and assigning each Gaussian a discrete index to this bank.
  This approach enables rich semantic feature association with minimal additional storage and computational overhead, thus preserving the rapid inference capabilities of feed-forward 3DGS.
  We demonstrate that SegSplat delivers geometric fidelity comparable to state-of-the-art feed-forward reconstruction methods while simultaneously enabling versatile open-set semantic querying, all without necessitating scene-specific optimization.
  Our work bridges a critical gap, paving the way for practical, on-the-fly generation of semantically rich 3D environments.
\end{abstract}


\section{Introduction}
\label{sec:intro}

The ability to reconstruct 3D environments with rich, open-set semantic understanding is increasingly critical for intelligent systems, enabling applications ranging from robotic navigation to immersive augmented reality. These scenarios demand not only accurate geometry but also the capacity to identify, query, and manipulate arbitrary high-level concepts -- such as "a specific type of chair", "any vehicle", or "all metallic objects" -- directly within the 3D representation~\cite{kerr2023lerf}.
While progress has been made in both 3D reconstruction~\cite{mildenhall2020nerf,yu2021pixelnerf,kerbl2023gaussiansplatting} and 2D open-set semantic segmentation~\cite{xu2023open,li2022language,liang2023open}, integrating these capabilities into a unified and efficient 3D representation that supports open-vocabulary queries without per-scene fine-tuning remains a fundamental challenge.

Recent advances in neural rendering, particularly Neural Radiance Fields (NeRFs)~\cite{mildenhall2020nerf}, have demonstrated compelling view synthesis through implicit scene representations. However, their reliance on per-scene optimization limits scalability and real-time deployment. In contrast, 3D Gaussian Splatting (3DGS)~\cite{kerbl2023gaussiansplatting} represents scenes as collections of 3D Gaussians, enabling high-quality, real-time rendering.
Despite these rendering advantages, training times remain substantial~\cite{kerbl2023gaussiansplatting}, motivating a shift towards faster feed-forward approaches~\cite{xu2024depthsplat,chen2024mvsplat,ye2024no}.

To bridge the semantic gap, recent works extend 3DGS with language-aligned features~\cite{qin2024langsplat,shi2023langan,ye2024gaussian}, typically associating each Gaussian with a feature vector (e.g., from CLIP~\cite{radford2021learning}). These enable open-vocabulary querying, but most methods still depend on per-scene optimization  (first constructing the 3DGS, then fitting semantic features), making them ill-suited for dynamic or large-scale environments.

This paper introduces SegSplat, a novel framework that predicts 3D Gaussian Splats enriched with open-set semantic features in a single feed-forward pass. Building on recent progress in sparse-view 3DGS reconstruction, we leverage the DepthSplat architecture~\cite{xu2024depthsplat} for fast and robust geometric prediction. Our key contribution is the seamless integration of semantic information, extracted via 2D open-set segmentation models (e.g., SAM~\cite{kirillov2023segment} with CLIP), directly into the predicted Gaussians.

To achieve this, we construct a compact semantic memory bank from features observed in the input views and assign a discrete index (analogous to a one-hot code) to each 3D Gaussian primitive. This allows each Gaussian to reference a powerful semantic descriptor while introducing minimal storage and computational overhead, preserving fast inference.

In summary, SegSplat provides a unified, efficient pipeline that jointly reconstructs 3D scene geometry and embeds open-vocabulary semantic features -- eliminating the need for per-scene optimization. We validate our method on challenging datasets including 3D-OVS~\cite{liu2023weakly} and RealEstate10k~\cite{realestate10k}, demonstrating that SegSplat achieves accurate feed-forward 3DGS quality while enabling robust open-set semantic querying.

\section{Related Work}
\label{sec:related_work}

Our work integrates progress in three key areas: 3D scene representations, their accelerated sparse feed-forward generation, and the incorporation of open-vocabulary semantic understanding.
This review situates SegSplat's contribution: a method for feed-forward generation of 3D Gaussian Splatting (3DGS) models with associated open-set semantic features.

The pursuit of high-fidelity 3D scene models have evolved from traditional representations like meshes, which often involve complex creation pipelines, towards learnable techniques.
Neural Radiance Fields (NeRF)~\cite{mildenhall2020nerf} were a significant advance, using neural networks to synthesize photorealistic novel views.
However, NeRFs typically require lengthy per-scene optimization, limiting their practical use.
3D Gaussian Splatting (3DGS)~\cite{kerbl2023gaussiansplatting} offered a compelling alternative, explicitly representing scenes with 3D Gaussian primitives.
This approach achieves state-of-the-art visual quality and real-time rendering via a differentiable rasterizer, making 3DGS a strong foundation for complex scene understanding tasks.
However, such approaches still require expensive per-scene optimizations.

To address the computational demands of this per-scene optimization, research has shifted towards feed-forward methods that generate 3D representations in a single feed-forward pass from a sparse input images.
These models are trained on large datasets to predict scene parameters directly. For 3DGS, methods like MVSplat~\cite{chen2024mvsplat}, DepthSplat~\cite{xu2024depthsplat}, and NoPoSplat~\cite{ye2024no} demonstrate rapid reconstruction by employing pipelines that typically involve 2D feature extraction, geometric inference (e.g., depth estimation), and regression of Gaussian attributes.
While these approaches efficiently reconstruct geometry and appearance, they do not inherently produce semantic information. Integrating semantics requires separate, post-hoc processing.

Incorporating semantic understanding into 3D models is crucial for higher-level reasoning. A common strategy is to "lift" features or predictions from powerful pre-trained 2D foundation models, such as CLIP~\cite{radford2021learning} for image-text understanding and SAM~\cite{kirillov2023segment} for class-agnostic segmentation. Projecting or distilling information from these 2D models into 3D representations is effective but presents challenges, including maintaining multi-view consistency and handling occlusions.

Several works have focused on creating 3D representations that store and query semantic information. For NeRFs, LERF~\cite{kerr2023lerf} enabled open-vocabulary querying by distilling CLIP features into a 3D language field, but this process requires per-scene optimization for the semantic component. Similar efforts for 3DGS, such as LangSplat~\cite{qin2024langsplat} and LEGaussians~\cite{shi2023langan}, associate language features with Gaussian primitives, also relying on per-scene optimization to embed these semantics.

The high dimensionality of language features stored within each Gaussian makes it computationally expensive for language-embedded 3DGS methods to rasterize them directly~\cite{qin2024langsplat}.
To address this issue various compression and quantization strategies have been proposed.
LangSplat adopts a pretrained scene-specific autoencoder~\cite{qin2024langsplat} to compress the language features into a lower-dimensional space used for rasterization.
After novel view rendering, the autoencoder is used to decode the language features to their original size.
LEGaussians uses quantized language feature index maps to avoid storing and rasterizing a full language feature for each Gaussian ~\cite{shi2023langan}.
A low-dimensional semantic feature vector is added to each gaussian which is then decoded into a discrete index in the feature index map using a small scene-specific MLP after the rasterization step.

Gaussian Grouping~\cite{ye2024gaussian} reparametrizes the open-set segmentation problem as a generic object segmentation task. It learns an identity encoding for each Gaussian representing object associations from video-tracked object masks found in the input images. Because each gaussian is not directly imbued with any open-set semantic features this approach requires a secondary step where a 2D object detection model such as GroundingDINO~\cite{liu2023grounding} is used to associate a text prompt with a 3D object in the scene.
While methods like LBG~\cite{chacko2025lifting} offer training-free mechanisms to add semantics to pre-existing 3DGS models, they depend on an already reconstructed geometric model and do not integrate semantic generation with geometric reconstruction.

Many existing methods for semantic 3D scene understanding, thus, face a bottleneck: even if geometry is reconstructed quickly, embedding rich semantic features often requires a separate, computationally intensive per-scene optimization step.
This limits their applicability in scenarios demanding rapid, on-the-fly generation of semantic 3D models.
SegSplat directly addresses this limitation. It proposes a framework to predict both 3D Gaussian Splat geometry and associated open-set semantic features in a single feed-forward pass from sparse image inputs tailored to the 3DGS representation produced by feed-forward gaussian splatting. This concurrent generation of geometry and semantics, without per-scene semantic optimization, is the key distinction of our approach.

\section{Proposed Method}
\label{sec:method}

\begin{figure}[t]
  \centering
  \includegraphics[width=0.95\linewidth]{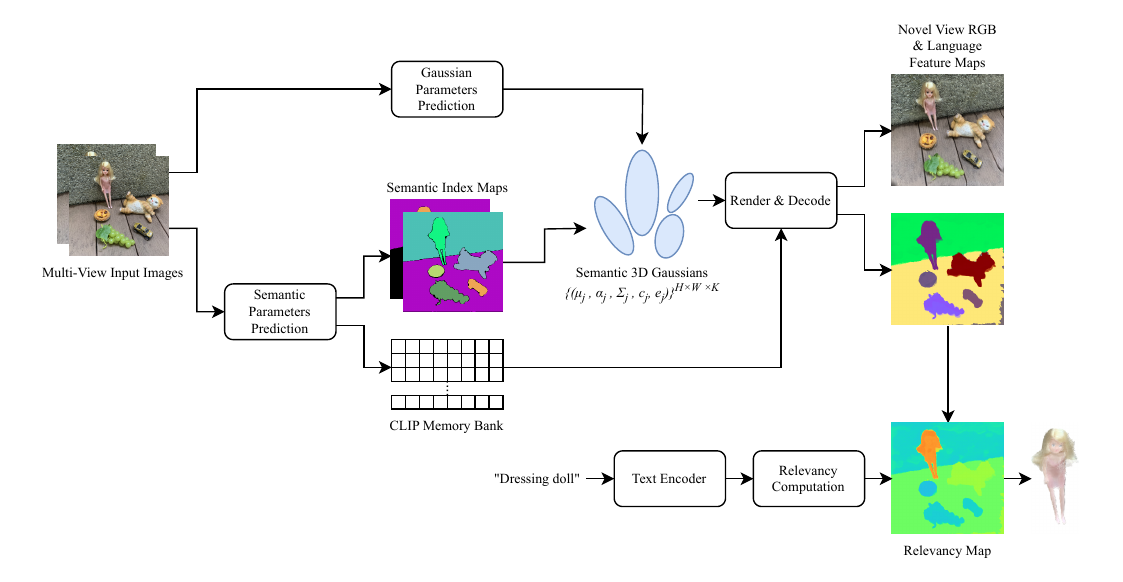}
  \caption{
    \textbf{SegSplat} predicts 3D Gaussian splats embedded with language features from sparse multi-view images, without any training.
    Our pipeline leverages pretrained DepthSplat to estimate 3D Gaussian parameters per pixel, and uses SAM+CLIP to extract segmentation masks and CLIP embeddings.
    To ensure memory efficiency, we construct a CLIP feature memory bank and represent per-object semantics using one-hot index maps aligned with this bank.
    These semantic indices are appended to the Gaussians predicted by DepthSplat.
    After splatting, we reconstruct full-length language features via an element-wise product between the rendered index maps and the memory bank.
    Novel-view querying is then performed on the decoded CLIP feature image.
  }
  \vspace{-5mm}
  \label{fig:segsplat_pipeline}
\end{figure}

Our proposed method, SegSplat, enables feed-forward generation of 3D Gaussian Splatting (3DGS) representations augmented with open-set semantic features from sparse multi-view images.
The core idea is to: (1) construct a semantic feature bank by leveraging Segment Anything Model (SAM) and CLIP extracted from input views, followed by clustering to identify representative semantic concepts.
(2) Utilize a feed-forward model (DepthSplat) to generate 3D Gaussian primitives, where each primitive is associated with a semantic index corresponding to an entry in our feature bank.
(3) Render both appearance and semantic features for novel views, enabling open-vocabulary querying.

\subsection{Semantic Parameter Prediction}
\label{subsec:semantic_parameters}

The aim of this component of the pipeline is to represent semantic information found in the input images with a memory bank of CLIP features and corresponding per-pixel semantic indices.
This memory bank will allow to have an efficient Gaussian-to-semantic assignment by simply assigning the index of the semantics from the bank to a given Gaussian primitive.
Given $K$ input images $\{\bm{I}_k \in \mathbb{R}^{H \times W \times 3}\}_{k=1}^{K}$ and their corresponding camera projection matrices $\{\bm{P}_k \in \mathbb{R}^{3 \times 4}\}_{k=1}^{K}$, we perform the following steps:
\begin{enumerate}
  \item \textbf{Mask Extraction}: For each input image $\bm{I}_k$, we employ SAM~\cite{kirillov2023segment} to generate a set of $N_k$ object masks $\{\bm{m}_{ki} \subset \mathbb{R}^2\}_{i=1}^{N_k}$. We prompt SAM at a ``whole'' object granularity and apply non-maximum suppression to ensure minimal overlap between masks. Each pixel $(u,v)$ in image $\bm{I}_k$ is thus associated with at most one mask.

  \item \textbf{CLIP Feature Extraction}: For each mask $\bm{m}_{ki}$, we create a cropped image patch. Pixels outside the mask $\bm{m}_{ki}$ within this crop are zeroed out. We then compute a $D_{\text{CLIP}}$-dimensional CLIP image embedding $\bm{f}_{ki} \in \mathbb{R}^{D_{\text{CLIP}}}$ using a pre-trained CLIP image encoder~\cite{radford2021learning}. This yields a collection of $N_{\text{total}} = \sum_{k=1}^K N_k$ mask-specific CLIP features.

  \item \textbf{Clustering and Bank Formation}: To create a concise semantic representation, we pool all $N_{\text{total}}$ CLIP features $\{\bm{f}_{ki}\}$. We then apply K-Means clustering to group these features into $M$ semantic clusters. The centroids of these $M$ clusters form our semantic feature bank $\bm{B} = [\bm{b}_1, \dots, \bm{b}_M]^T \in \mathbb{R}^{M \times D_{\text{CLIP}}}$ used in later processes. Each original CLIP feature $\bm{f}_{ki}$ is assigned to its closest cluster centroid $\bm{b}_m$.

  \item \textbf{Per-Pixel Semantic Index Maps}: Using the bank, we generate a semantic index map $\bm{S}_k \in \{1, \dots, M\}^{H \times W}$ for each input image $\bm{I}_k$. For a pixel $(u,v)$ in $\bm{I}_k$ belonging to mask $\bm{m}_{ki}$ (assigned to cluster $m$), we set $\bm{S}_k(u,v) = m$. Pixels not belonging to any mask are be assigned a background index $0$.
    This results in $K$ semantic index maps $\{\bm{S}_k\}_{k=1}^K$.
\end{enumerate}

\subsection{Feed-Forward Generation of Semantic Gaussians}
\label{subsec:gaussian_generation}

We adopt a feed-forward Gaussian Splatting architecture, specifically a frozen DepthSplat model~\cite{xu2024depthsplat}, to generate the 3D scene representation.
Beginning with $K$ sparse input images $\{\bm{I}_i\}_{i=1}^{K}, (\bm{I}^i \in \mathbb{R}^{H \times W \times 3})$, and their corresponding camera projection matrices $\{\bm{P}_i\}_{i=1}^{K}, (\bm{P}^i \in \mathbb{R}^{3 \times 4})$, DepthSplat predicts the per-pixel Gaussian parameters $\{(\bm{\mu}_j, \alpha_j, \bm{\Sigma}_j, \bm{c}_j)\}_{j=1}^{H \times W \times K}$ for each image. Gaussian parameters  $\bm{\mu}_j$, $\alpha_j$, $\bm{\Sigma}_j$, and $\bm{c}_j$ denote the 3D Gaussians position, opacity, covariance, and color (represented as spherical harmonics), respectively.

Because DepthSplat predicts per-pixel Gaussian parameters for each pixel in each input view, we can directly append a computed one-hot semantic encoding $\bm{e}_j$ derived from the semantic index maps calculated in Section \ref{subsec:semantic_parameters} to the other Gaussian parameters. Our full Gaussian splatting representation becomes $\bm{g}_j = \{(\bm{\mu}_j, \alpha_j, \bm{\Sigma}_j, \bm{c}_j, \bm{e}_j)\} \in \mathbb{R}_{j=1}^{H \times W \times K}$.
Such a representation allows for the light-weight assignment of semantics to Gaussians without significantly inflating the number of parameters that need to be stored for each primitive.
By the end of this step, each Gaussian has a corresponding semantic class assigned.

\subsection{Differentiable Rendering of Color and Semantics}
\label{subsec:rendering}

To synthesize a novel view given a camera pose $\bm{P}_{\text{novel}}$, we use the standard tile-based 3D Gaussian splatting rasterization process~\cite{kerbl2023gaussiansplatting}. For each pixel $v$ in the novel view it is the following.

\textbf{Color Rendering}:
For each pixel $v$, the color $C(v) \in \mathbb{R}^3$ is rendered as follows:
\begin{equation}
  C(v) = \sum_{i \in \mathcal{N}} c_i \alpha_i \prod_{j=1}^{i-1} (1 - \alpha_j),
  \label{eq:rendering_3dgs_color}
\end{equation}
where $c_i$ denotes the color of the $i$-th Gaussian, $\mathcal{N}$ is the set of Gaussians within a tile, and $\alpha_i = o_i G^{2D}_i(v)$, with $o_i$ being the opacity of the $i$-th Gaussian and $G^{2D}_i(\cdot)$ denoting the projection of the $i$-th Gaussian onto 2D.

\textbf{Semantic Index Rendering}:
Each 3D Gaussian $G_j$ in our scene representation is associated with a one-hot semantic vector $\bm{e}_i \in \{0,1\}^M$, indicating its assignment to one of the $M$ semantic concepts derived from our feature bank. To render the semantic information at a pixel in a novel view, we adapt the splatting mechanism used for color as follows:
\begin{equation}
  \bm{E}(v) = \sum_{i \in \mathcal{N}} \bm{e}_i \alpha_i \prod_{j=1}^{i-1} (1 - \alpha_j),
  \label{eq:rendering_3dgs_semantics}
\end{equation}
where $\bm{E}(v)$ represents the semantic embedding rendered at pixel $v$.
This vector $\bm{E}(v)$ represents a blended distribution of semantic concepts at the pixel, where each element $(\bm{E}(v))_m$ indicates the aggregated presence of the $m$-th semantic concept.

\textbf{CLIP Feature Map Recovery}:
After rendering, each pixel possesses a blended semantic vector $\bm{E}(v) \in \mathbb{R}^M$. Our semantic feature bank $\bm{B} \in \mathbb{R}^{M \times D_{\text{CLIP}}}$ stores the $D_{\text{CLIP}}$-dimensional CLIP embeddings for the $M$ representative semantic concepts (i.e., the $m$-th row of $\bm{B}$, denoted $\bm{b}_m^T$, is the CLIP feature vector for the $m$-th semantic concept).
To recover a continuous CLIP feature vector $\bm{F}(v) \in \mathbb{R}^{D_{\text{CLIP}}}$ for the pixel, we use $\bm{E}(v)$ to linearly combine the feature vectors stored in the bank $\bm{B}$.
Specifically, the $m$-th component of $\bm{E}(v)$, $(\bm{E}(v))_m$, acts as the weight for the $m$-th semantic feature vector $\bm{b}_m$ as follows:
\begin{equation}
  \bm{F}(v) = \sum_{m=1}^{M} (\bm{E}(v))_m \cdot \bm{b}_m = \bm{E}(v)^T \bm{B}.
  \label{eq:clip_recovery_detailed} 
\end{equation}
This operation effectively translates the rendered (potentially mixed) semantic indices back into the rich, continuous CLIP feature space. The resulting $\bm{F}(v)$ for all pixels forms a dense CLIP feature map for the novel view, which can then be used for open-vocabulary tasks.

As a result, the contributions of each CLIP feature in the memory bank is weighted by the rendered mask encoding.
Finally, before performing open-vocabulary querying, the CLIP features are normalized to unit length.

\subsection{Open-Vocabulary Querying}
\label{subsec:querying}

With the rendered CLIP feature map $\bm{F}(v)$, we can perform open-vocabulary queries. Given a text query $q_{\text{text}}$, we compute its CLIP text embedding $\phi_{qry} \in \mathbb{R}^{D_{\text{CLIP}}}$. For each pixel with rendered CLIP image feature $\phi_{\text{img}} = \bm{F}(v)$, we calculate a relevancy score following LERF~\cite{kerr2023lerf} as:
\begin{equation}
  \text{relevancy}(\phi_{\text{img}}, \phi_{qry}) = \min_{i \in \{\text{obj, thg, stf}\}} \left( \frac{\exp(\tau \cdot \phi_{\text{img}} \cdot \phi_{qry})}{\exp(\tau \cdot \phi_{\text{img}} \cdot \phi_{qry}) + \exp(\tau \cdot \phi_{\text{img}} \cdot \phi_{i_{\text{canon}}})} \right)
  \label{eq:relevancy}
\end{equation}
where $\phi_{i_{\text{canon}}}$ are CLIP text embeddings of predefined canonical phrases like ``object'', ``things'', ``stuff'', and $\tau$ is a temperature parameter.
Relevancy scores below a predefined threshold (e.g., 0.5) are set to 0, and the remaining scores can be further thresholded to obtain fine-grained segmentation masks corresponding to the input query.

\section{Experiments}
\label{sec:experiments}

In this section, we evaluate our method on standard benchmarks for 3D open-set semantic segmentation, and we also demonstrate that SegSplat achieves the same photometric scores as the state-of-the-art DepthSplat.

\textbf{Datasets.}
We evaluate our method on two datasets: RealEstate10K (RE10k)~\citep{realestate10k} and 3D-OVS~\citep{liu2023weakly}.
RE10k comprises a large collection of real estate videos with estimated camera parameters. For RE10k, we use an official subset and follow the train/test split of DepthSplat~\citep{xu2024depthsplat}. As RE10k lacks mask-level annotations, we present qualitative results for semantic segmentation.

3D-OVS~\citep{liu2023weakly} features scenes with long-tailed object distributions in diverse backgrounds. For sparse multi-view reconstruction on 3D-OVS, we adopt the train/test split generation methodology from MVSplat~\citep{chen2024mvsplat} and DepthSplat~\citep{xu2024depthsplat}. Our test split uses input views with at least 60\% projected overlap; target views are those containing ground-truth semantic labels. We report Intersection over Union (IoU) for 3D semantic segmentation on 3D-OVS.

\textbf{Baselines.}
We compare SegSplat against several state-of-the-art methods. It is crucial to distinguish between methods that perform per-scene optimization or require access to the \emph{entire scene context} for semantic understanding, and feed-forward approaches like ours that operate on a \emph{limited set of input views} (typically two in our experiments).

First, we consider baselines that leverage extensive information from the target scene, often through per-scene training or optimization.
Consequently, they are expected to achieve higher accuracy and primarily serve as a reference for the semantic segmentation task, rather than direct competitors to our feed-forward approach.
This group includes open-vocabulary 2D segmentation models such as LSeg~\citep{li2022language}, ODISE~\citep{xu2023open}, and OV-Seg~\citep{liang2023open}, for which metrics are typically derived from individual views without enforcing 3D multi-view consistency.
We also include NeRF-based methods like FFD~\citep{kobayashi2022decomposing}, LERF~\citep{kerr2023lerf}, and 3D-OVS~\citep{liu2023weakly}, requiring per-scene optimization. Furthermore, non-feed-forward Gaussian Splatting semantic methods such as GS-Grouping~\citep{ye2024gaussian}, and  LangSplat~\citep{qin2024langsplat} fall into this category, as they typically operate on or optimize semantic features for a 3DGS scene representation, benefiting from full scene context.

Second, to ensure a fair comparison for our feed-forward SegSplat, we create a feed-forward comparison group.
For this, we adapt the state-of-the-art LangSplat~\citep{qin2024langsplat} to operate within the same feed-forward, sparse-view framework as SegSplat.
Instead of using the pre-trained or optimized 3DGS representations from their original implementations, LangSplat is initialized using the exact same Gaussians predicted by our common frozen DepthSplat backbone from only two input views.
All methods in this group utilize identical semantic features extracted from SAM (masks) and CLIP (embeddings) as input.
LangSplat requires a scene-specific autoencoder. We train it for each scene individually (2k iterations, batch size 2, learning rate $2.0 \times 10^{-2}$, AdamW optimizer).
The time taken for this per-scene autoencoder training is factored into LangSplat's reported run-time to ensure a fair comparison against methods that do not require scene-specific training.

\textbf{Implementation details.}
Our pipeline utilizes the 37M parameter DepthSplat model, pre-trained on RE10k with 256$\times$256 resolution images, to predict initial Gaussian parameters.
The DepthSplat model is not fine-tuned on 3D-OVS, allowing us to evaluate the zero-shot generalization of our semantic assignment approach.
To ensure view-invariant semantic rendering, the spherical harmonics degree for the semantic index encoding is set to zero during rasterization.
We replace DepthSplat's original rasterizer~\citep{kerbl2023gaussiansplatting} with \texttt{gsplat}~\citep{ye2025gsplat} for rendering both color and our semantic index encodings.

For mask extraction from input images (Section~\ref{subsec:semantic_parameters}), we employ the SAM ViT-H model~\citep{kirillov2023segment}. We use the CLIP ViT-B/16 model~\citep{radford2021learning}, trained on the LAION-2B English subset of LAION-5B~\citep{schuhmann2022laionb}, to extract features from SAM-identified objects and to encode text prompts.
The number of clusters M for K-Means during semantic feature bank construction (Section~\ref{subsec:semantic_parameters}) is $M = \lambda N_\text{total} / K$ where $N_\text{total}$ is the total number of masks from all $K$ input views, and $\lambda = 1.2$. This heuristic aims to accommodate varying object visibility across views.
For open-vocabulary querying (Section~\ref{subsec:querying}), relevancy scores are computed as in LERF~\citep{kerr2023lerf}.
Scores below 0.5 are set to 0. The resulting relevancy map is then thresholded at 0.5 to produce binary segmentation masks.

All experiments are conducted on an NVIDIA RTX-5090 GPU, and we report inference/processing times. The primary evaluation of SegSplat's performance and efficiency should be made against other methods within the "Feed-Forward GS", as they operate under similar constraints.

\subsection{Results}

\begin{table}[t]
  \centering
  \caption{\textbf{3D semantic segmentation} performance comparison (mIoU \%) on the 3D-OVS dataset~\citep{liu2023weakly}. Methods are grouped by their operational principle. Best results within the feed-forward category (run only on two input views) and leading results among contextual (non-feed-forward; trained on \textit{all} images from a scene) methods are shown in \textbf{bold}.}
  \resizebox{\textwidth}{!}{\begin{tabular}{l l c c c c c c c}
    \toprule
    \textbf{Principle} & \textbf{Method} & \textbf{Bed} & \textbf{Bench} &
    \textbf{Room} &
    \textbf{Sofa} & \textbf{Lawn} & \textbf{Overall} \\
    \midrule
    \multirow{3}{*}{2D Image Segmentation}
    & LSeg \citep{li2022language} & 56.0 & \phantom{1}6.0  & 19.2 & \phantom{1}4.5  & 17.5 & 20.6 \\
    & ODISE \citep{xu2023open} & 52.6 & 24.1 & 52.5 & 48.3 & 39.8 & 43.5 \\
    & OV-Seg \citep{liang2023open} & 79.8 & 88.9 & 71.4 & 66.1 & 81.2 & 77.5 \\
    \multirow{3}{*}{NeRF}
    & FFD \citep{kobayashi2022decomposing}             & 56.6 &
    \phantom{1}6.1  &
    25.1 & \phantom{1}3.7  & 42.9 & 26.9 \\
    & LERF \citep{kerr2023lerf}           & 73.5 & 53.2 & 46.6 & 27.0
    & 73.7 & 54.8 \\
    & 3D-OVS    \citep{liu2023weakly}      & 89.5 & 89.3 & 92.8 & 74.0
    & 88.2 & 86.8 \\
    \multirow{2}{*}{Gaussian Splatting}

    & GS-Grouping \citep{ye2024gaussian}    & 83.0 & 91.5 & 85.9 &
    87.3 & 90.6 & 87.7 \\
    & LangSplat \citep{qin2024langsplat}      &  \textbf{92.5} &
    {\textbf{94.2}} &
    {\textbf{94.1}} & \textbf{90.0} & {\textbf{96.1}} & {\textbf{93.4}} \\
    \midrule
    \multirow{2}{*}{Feed-Forward GS}
    & LangSplat \citep{qin2024langsplat} & 59.4 & 75.1 & \phantom{1}7.2 & 50.9 & 46.0 & 47.7 \\
    & \textbf{SegSplat} & \textbf{66.2} & \textbf{75.9} & \textbf{19.8} & \textbf{64.5} & \textbf{80.8} & \textbf{61.4} \\
    \bottomrule
  \end{tabular}}
  \label{table:iou_comparison}
\end{table}

\begin{table}[t]
  \centering
  \caption{\textbf{Novel view synthesis quality} for SegSplat and its base geometric model, DepthSplat, on the RE10k and 3D-OVS datasets. Metrics reported are PSNR$\uparrow$, SSIM$\uparrow$, and LPIPS$\downarrow$. The identical performance demonstrates that SegSplat's method of integrating semantic features does not degrade the geometric and appearance reconstruction fidelity of the underlying DepthSplat model, as SegSplat utilizes the same Gaussian parameters for rendering.}
  \begin{tabular}{l l c c c}
    \toprule
    \textbf{Dataset} & \textbf{Method} & \textbf{PSNR $\uparrow$} & \textbf{SSIM $\uparrow$} &
    \textbf{LPIPS $\downarrow$} \\
    \midrule
    \multirow{2}{*}{3D-OVS~\citep{liu2023weakly}} & DepthSplat~\citep{xu2024depthsplat} & 16.99 & 0.392 & 0.428 \\
    & \textbf{SegSplat} & 16.99 & 0.392 & 0.428 \\
    \midrule
    \multirow{2}{*}{RE10K~\citep{realestate10k}} & DepthSplat~\citep{xu2024depthsplat} & 25.89 & 0.881 & 0.122 \\
    & \textbf{SegSplat} & 25.89 & 0.881 & 0.122 \\
    \bottomrule
  \end{tabular}
  \vspace{-5mm}
  \label{table:segsplat_gaussian_metrics}
\end{table}

\begin{figure}[ht]
  \centering
  \includegraphics[width=0.99\linewidth]{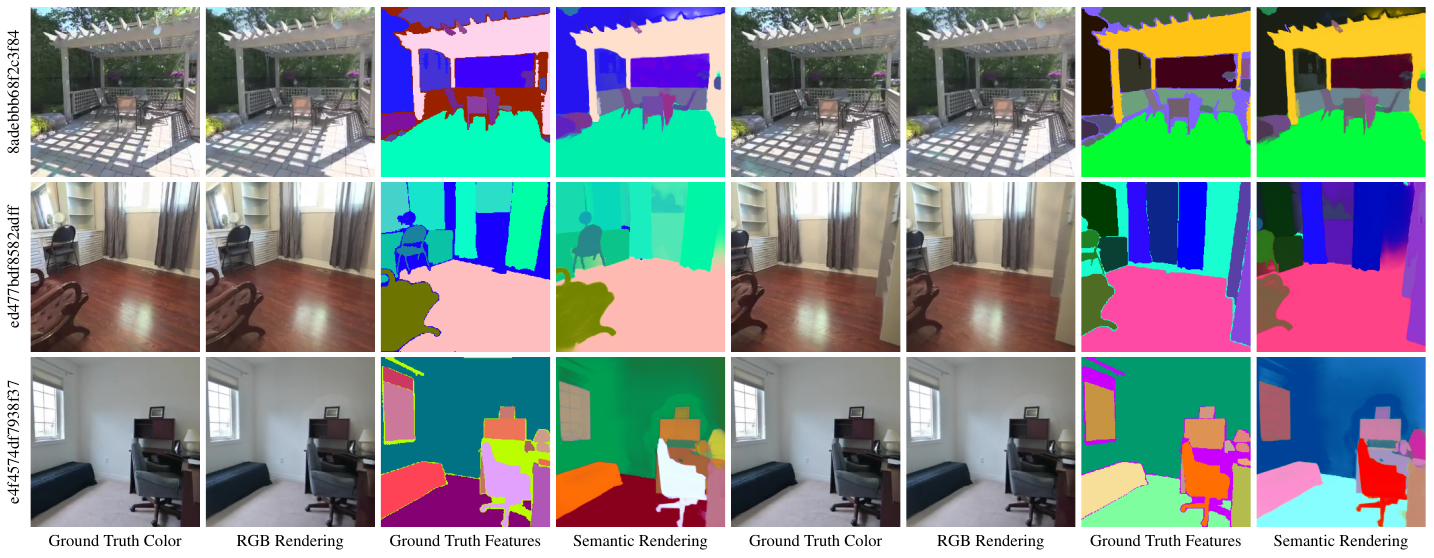}
  \caption{
    Comparison of predicted and ground truth (GT) color and semantic maps for novel views rendered by SegSplat on the RealEstate10K dataset~\cite{realestate10k}.
    Semantic maps are visualized using PCA. Ground truth semantic features are obtained by applying SAM+CLIP to the corresponding GT novel view images.
    Each group of four columns shows: GT RGB image, SegSplat-rendered RGB, GT semantics, and SegSplat-rendered semantics. This sequence is repeated for a second novel view.
  }
  \label{fig:re10k_segsplat_scene_comparison}
\end{figure}

\begin{figure}[htp]
  \centering
  \includegraphics[width=0.99\linewidth]{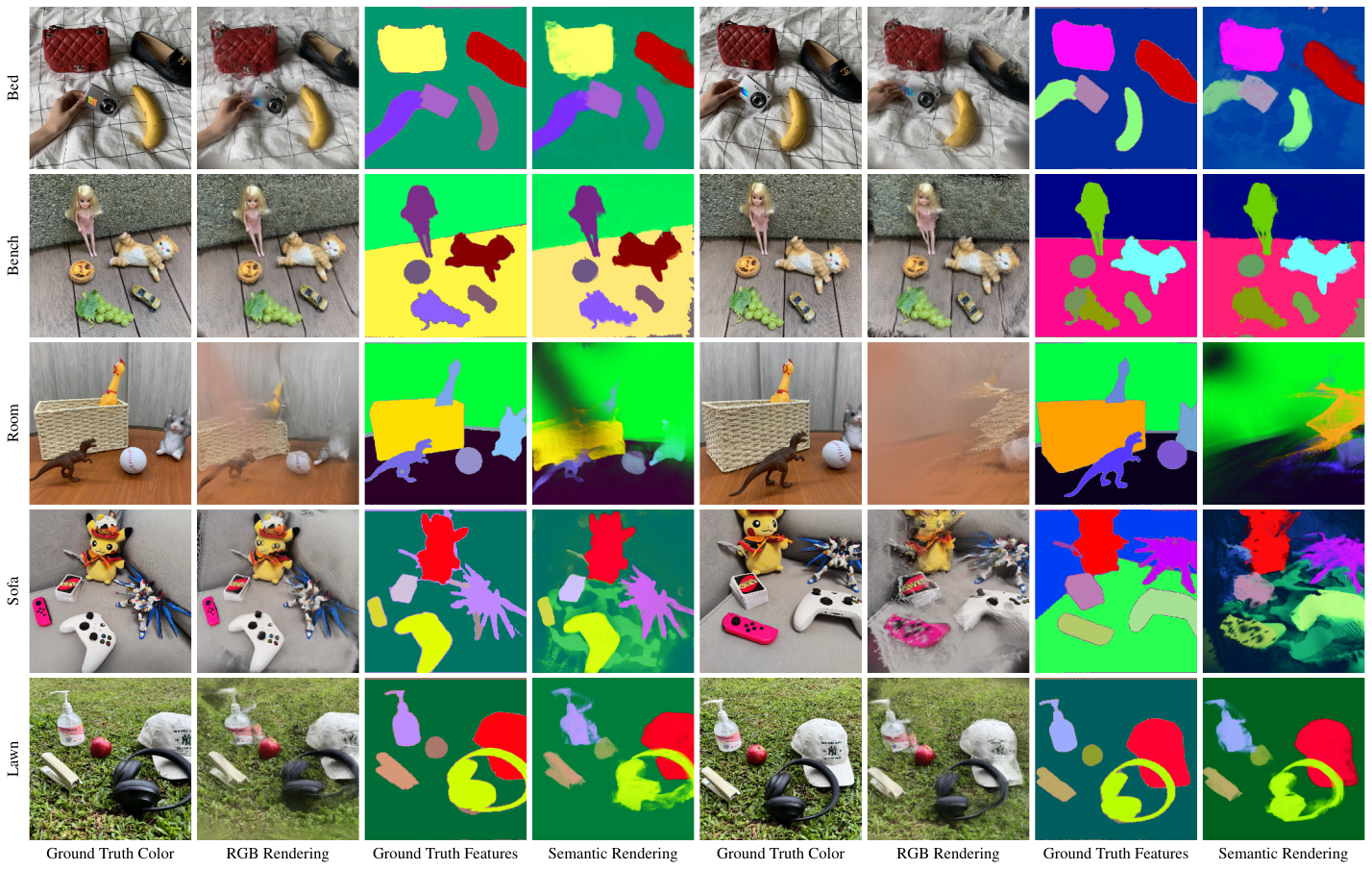}
  \caption{
    Comparison of predicted and ground truth (GT) color and semantic maps for novel views rendered by SegSplat on the 3D-OVS dataset~\cite{liu2023weakly}.
    Semantic maps are visualized using PCA. Ground truth semantic features are obtained by applying SAM+CLIP to the corresponding GT novel view images.
  Each group of four columns shows: GT RGB image, SegSplat-rendered RGB, GT semantics, and SegSplat-rendered semantics. This sequence is repeated for a second novel view.}
  \label{fig:3d_ovs_segsplat_scene_comparison}
\end{figure}

\begin{figure}[t]
  \centering
  \subfloat[\centering ``Bench'' Scene open-vocabulary segmentation.\label{fig:bench_segmentation}]{
    \includegraphics[height=5cm]{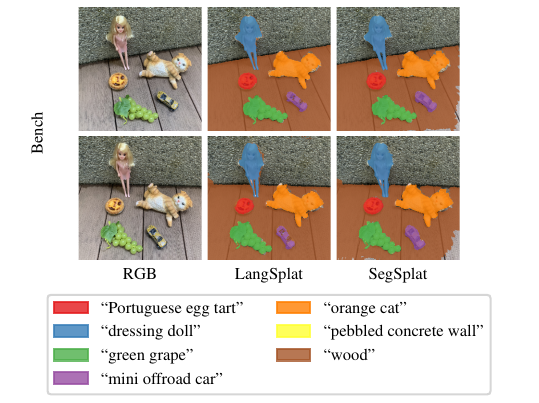}
  }%
  \subfloat[\centering ``Lawn'' Scene open-vocabulary segmentation.\label{fig:lawn_segmentation}]{
    \includegraphics[height=5cm]{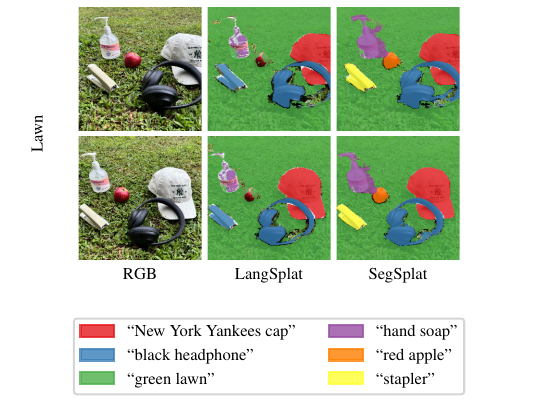}
  }%
  \caption{A qualitative comparison of the masks produced by SegSplat and LangSplat on the 3D-OVS dataset~\cite{liu2023weakly}. We show results for two scenes and two different novel views. We observe that our method produces more accurate segmentation masks.}
  \label{fig:3d_ovs_segsplat_segmentation_comparison}
\end{figure}

\textbf{3D Semantic Segmentation Performance.}
We evaluated open-vocabulary 3D semantic segmentation capabilities of the proposed SegSplat on the 3D-OVS dataset, with results presented in Table~\ref{table:iou_comparison}.
The methods are benchmarked using mean Intersection over Union (mIoU) and are categorized based on their underlying methodology.
Examples are shown in Figs.~\ref{fig:re10k_segsplat_scene_comparison}, \ref{fig:3d_ovs_segsplat_scene_comparison} and \ref{fig:3d_ovs_segsplat_segmentation_comparison}.

The table includes several baselines that perform per-scene optimization or leverage extensive information from the target scene.
These encompass 2D image segmentation techniques (LSeg, ODISE, OV-Seg), NeRF-based semantic methods (FFD, LERF, 3D-OVS), and Gaussian Splatting approaches that optimize semantics for the entire scene (GS-Grouping, LangSplat). These methods, such as the original LangSplat which achieves an overall mIoU of 93.4\%, provide a strong performance reference due to their comprehensive access to scene data.

Our primary evaluation focuses on the feed-forward Gaussian Splatting (Feed-Forward GS) category, which operates under the constraint of using only a few input images (two in our setup) without any per-scene optimization for semantics.
Within this challenging setting, SegSplat achieves an overall mIoU of 61.4\%.
This significantly surpasses the adapted feed-forward version of LangSplat, which scores 47.7\% mIoU when constrained to the same feed-forward pipeline using an identical DepthSplat backbone and semantic inputs.
SegSplat demonstrates superior performance across the majority of reported object categories in this direct comparison, underscoring its efficacy in generating robust semantic segmentations in a single pass.

While a performance difference exists between feed-forward approaches like SegSplat and methods that utilize full scene context and optimization, SegSplat establishes a strong baseline for efficient, open-set semantic understanding with 3D Gaussian Splatting in a purely feed-forward manner.
Moreover, it does not require additional per-scene training as LangSplat does.

Table \ref{fig:scannet} reports open-set 3D semantic segmentation results on the ScanNet++ dataset~\citep{yeshwanth2023scannet++}. SegSplat achieves the highest mean IoU among feed-forward approaches. We compare against the recent SAB3R variants~\citep{chen2025sab3r} and LSM~\citep{fan2024large}. SegSplat with SAM attains 26.0 percent mIoU and surpasses prior feed-forward baselines by a clear margin. EfficientSAM reduces preprocessing time from 9.70 seconds to 0.98 seconds with a moderate decrease in accuracy, which shows that SegSplat remains effective when using faster mask extraction. Replacing the DepthSplat backbone with MVSplat yields 22.2 percent mIoU, demonstrating that SegSplat generalizes across feed-forward Gaussian Splatting backbones. Inference remains 0.03 seconds for all SegSplat variants.

\begin{table}[t]
\centering
\caption{Open-set 3D semantic segmentation results on ScanNet++~\citep{yeshwanth2023scannet++}. SegSplat achieves the highest mean IoU among feed-forward methods while retaining low inference time. We report results for SegSplat using EfficientSAM~\citep{xiong2024efficientsam} and also using an MVSplat~\citep{chen2024mvsplat} backbone. Preprocessing time corresponds to SAM-based mask extraction.}
\resizebox{\textwidth}{!}{\begin{tabular}{lcccc}
\hline
Method & mean IoU (\%) & Training & Preprocessing & Inference \\
\hline
SAB3R (B)~\citep{chen2025sab3r} & \phantom{1}4.6 & not reported & -- & $\leq$0.108 s \\
SAB3R (C)~\citep{chen2025sab3r} & 17.3 & not reported & -- & $\leq$0.108 s \\
SAB3R (CD)~\citep{chen2025sab3r} & 17.5 & not reported & -- & $\leq$0.108 s \\
LSM~\citep{fan2024large} & 21.4 & 3 days on 8$\times$A100 & -- & 0.108 s \\
SegSplat (SAM) & \textbf{26.0} & -- & 9.70 s & 0.030 s \\
\hline
SegSplat (E-SAM~\citep{xiong2024efficientsam}) & 21.2 & -- & 0.98 s & 0.030 s \\
SegSplat w/ MVSplat & 22.2 & -- & 9.70 s & 0.030 s \\
\hline
\end{tabular}}
\label{fig:scannet}
\end{table}

\textbf{Geometric Reconstruction Quality.}
We evaluated whether the integration of our semantic component affects the underlying novel view synthesis quality.
SegSplat utilizes the geometric and appearance parameters for its Gaussians directly from the frozen DepthSplat model.
The primary goal of this comparison is to demonstrate that our method for adding semantic features preserves the rendering fidelity of the base model.

Table~\ref{table:segsplat_gaussian_metrics} presents standard image quality metrics (PSNR, SSIM and LPIPS) for SegSplat and baseline DepthSplat in the 3D-OVS and RE10k datasets. As shown, SegSplat achieves identical performance to DepthSplat across all metrics on both datasets. This confirms that our approach for incorporating open-set semantic understanding does not introduce any degradation to the high-fidelity geometric and appearance reconstruction capabilities provided by the underlying feed-forward 3DGS model.

\begin{table}[t]
  \centering
  \caption{
    \textbf{Runtime breakdown of SegSplat and LangSplat (in seconds).}
    Per-component runtime on the 3D-OVS dataset measured on a single RTX 5090 GPU and averaged after two warm-up batches.
    LangSplat needs scene-specific training (over 9 mins per scene) before inference, while SegSplat performs all steps in a single feed-forward pass.
    After SAM+CLIP features are extracted, SegSplat maintains a total inference time under 0.2 seconds, making it suitable for real-time tasks.
  }
  \begin{tabular}{lcr}
    \toprule
    \textbf{Component} & \textbf{SegSplat (s)} & \textbf{LangSplat (s)} \\
    \midrule
    2D Semantic Feature Extraction & 9.7000 & 9.7000 \\
    Semantic Gaussians Prediction  & 0.0284 & 565.0236 \\
    Gaussian Rasterization      & 0.0013 & 0.0016 \\
    Language Feature Decoder    & 0.0004 & 0.0001 \\
    Text Querying               & 0.0003 & 0.0002 \\
    \midrule
    \textbf{Total}              & \textbf{9.7304} & \textbf{575.1255} \\
    \bottomrule
  \end{tabular}
  \label{table:runtime_comparison}
  \vspace{-5mm}
\end{table}

\textbf{Processing Time.}
Table~\ref{table:runtime_comparison} presents a detailed runtime comparison of SegSplat and LangSplat, measured on a single RTX 5090 GPU.
While both methods require initial 2D semantic feature extraction using SAM+CLIP -- which accounts for the majority of the offline cost -- SegSplat eliminates the need for scene-specific optimization by performing semantic-to-Gaussian assignment in a single feed-forward pass.
In contrast, LangSplat relies on an expensive 565-second scene-specific training stage for semantic association.
Excluding shared preprocessing, SegSplat achieves a total inference-time runtime of under 0.2 seconds, over three orders of magnitude faster than LangSplat.
This efficiency makes SegSplat highly suitable for real-time or on-the-fly 3D scene segmentation applications, particularly in dynamic or large-scale environments.

\textbf{Limitations.}
Despite its advancements, SegSplat has limitations defining future research directions. Its semantic understanding quality is tied to the performance of the underlying 2D foundation models (SAM and CLIP) and the K-Means clustering used for its semantic bank. A key challenge is our current reliance on aggregating per-image 2D semantic predictions without a learned 3D fusion mechanism to optimally handle view inconsistencies or occlusions during the 2D-to-3D lifting process.
Additionally, geometric accuracy is inherited from the frozen DepthSplat backbone, and the system currently focuses on static scenes.
Consequently, while SegSplat offers significant efficiency with sparse inputs, its semantic segmentation accuracy lags behind methods that perform per-scene optimization using more extensive view information, a trade-off inherent to its feed-forward design.

\section{Conclusion}
\label{sec:conclusion}

We have introduced SegSplat, a novel framework designed to bridge the gap between rapid, feed-forward 3D reconstruction and rich, open-vocabulary semantic understanding.
By constructing a compact semantic memory bank from multi-view 2D foundation model features and predicting discrete semantic indices alongside geometric and appearance attributes for each 3D Gaussian in a single pass, SegSplat efficiently imbues scenes with queryable semantics.
Our experiments demonstrate that SegSplat achieves geometric fidelity comparable to state-of-the-art feed-forward 3D Gaussian Splatting methods while simultaneously enabling robust open-set semantic segmentation, crucially \textit{without} requiring any per-scene optimization for semantic feature integration.
This work represents a significant step towards practical, on-the-fly generation of semantically aware 3D environments, vital for advancing robotic interaction, augmented reality, and other intelligent systems.





\bibliography{iclr2026_conference}
\bibliographystyle{iclr2026_conference}

\appendix
\section{Appendix}
You may include other additional sections here.

\end{document}